\documentclass[10pt, a4paper]{article}

\usepackage{lrec-coling2024} 

\usepackage{times}
\usepackage{latexsym}

\usepackage[T1]{fontenc}

\usepackage[utf8]{inputenc}

\usepackage{microtype}

\usepackage{inconsolata}
\usepackage[]{linguex}
\usepackage{CJKutf8}
\usepackage{booktabs}
\usepackage{multirow}
\usepackage{graphicx}

\newcommand{\bert}{\texttt{BERT}\xspace}
\newcommand{\roberta}{\texttt{RoBERTa}\xspace}
\newcommand{\bertwwm}{\texttt{BERT-wwm}\xspace}
\newcommand{\mbert}{\texttt{mBERT}\xspace}
\newcommand{\bilstm}{\texttt{BiLSTM}\xspace}

\title{Computational Modelling of Plurality and Definiteness\\ in Chinese Noun Phrases}

\name{Yuqi Liu$^{\spadesuit}$, Guanyi Chen$^{\heartsuit\dag}$\thanks{$^\dag$Corresponding Author}, Kees van Deemter$^{\spadesuit}$} 

\address{$^{\heartsuit}$Hubei Provincial Key Laboratory of Artificial Intelligence and Smart Learning, \\
  National Language Resources Monitoring and Research Center for Network Media, \\
  School of Computer Science, Central China Normal University \\
$^{\spadesuit}$Department of Information and Computing Sciences, Utrecht University\\
         \texttt{y.liu30@students.uu.nl, g.chen@ccnu.edu.cn, c.j.vandeemter@uu.nl}}

\abstract{
Theoretical linguists have suggested that some languages (e.g., Chinese and Japanese) are ``cooler'' than other languages based on the observation that the intended meaning of phrases in these languages depends more on their contexts. As a result, many expressions in these languages are shortened, and their meaning is inferred from the context. In this paper, we focus on the omission of the plurality and definiteness markers in Chinese noun phrases (NPs) to investigate the predictability of their intended meaning given the contexts. To this end, we built a corpus of Chinese NPs, each of which is accompanied by its corresponding context, and by labels indicating its singularity/plurality and definiteness/indefiniteness. We carried out corpus assessments and analyses. The results suggest that Chinese speakers indeed drop plurality and definiteness markers very frequently. Building on the corpus, we train a bank of computational models using both classic machine learning models and state-of-the-art pre-trained language models to predict the plurality and definiteness of each NP. We report on the performance of these models and analyse their behaviours. The code and data used in this paper are available at: \url{https://github.com/andyzxq/chinese_np_def}.
\\ \newline \Keywords{Chinese Linguistics, Noun Phrase, Plurality, Definiteness}}

\begin{document}
\begin{CJK}{UTF8}{gbsn}
\maketitleabstract

\section{Introduction} \label{sec:intro}

It has been pointed out that speakers trade-off clarity against brevity~\citep{grice1975logic} and speakers of different languages appear to handle this trade-off differently~\citep{newnham1971about}. \citet{ross-1982-generating} and \citet{huang1984distribution} elaborated this idea by hypothesising that some languages (especially, Eastern Asian languages, e.g., Chinese and Japanese) are ``cooler'' than other languages. A language $A$ is considered to be cooler than language $B$ if understanding sentences of $A$ tends to require more work by readers or the hearers than understanding sentences of $B$. 
As a consequence, speakers of relatively cool languages often omit pronouns (causing \emph{pro-drop}) and assume that listeners can infer the missing information from the context.
Later on, the theory was extended, suggesting that many components in cool language are omittable~\citep{van1998adverbial}, such as plurality markers, definiteness markers~\citep{huang2009syntax}, discourse connectives~\citep{yu1993chinese} and so on. 

So far, most works have analysed related language phenomena as built into a language's grammar (e.g., the grammar of Chinese permits pro-drop). Only a few studies focused on the pragmatic aspects of coolness~\citep{chen-van-deemter-2020-lessons,chen-van-deemter-2022-understanding,chen2022computational}. For instance, \citet{chen-etal-2018-modelling} investigated the use of pro-drop by modelling the choices of speakers computationally. To the best of our knowledge, no similar study has focused on listeners' understanding. 

To fill this gap, we investigate the comprehension of two kinds of omittable information in Chinese noun phrases (NPs)\footnote{Note that we concentrate on Mandarin Chinese in this study.}, namely, plurality and definiteness, which are two major foci of research on NPs~\citep{iljic1994quantification, bremmers2022translation}.
The corresponding comprehension tasks for English are trivial because the plurality and definiteness of an English NP are always conveyed through explicit markers. In contrast, in Chinese, a bare noun can be either definite or indefinite and either singular or plural. Consider the following examples of the noun ``狗'' (\emph{dog}) from~\citet{huang2009syntax}:
\ex. \a. 狗 很 聪明 。\label{ex:1a}
\glt gou hen congming .
\glt `Dogs are intelligent.'
\b. 我 看到 狗 。\label{ex:1b}
\glt wo kandao gou .
\glt `I saw a dog/dogs.'
\c. 狗 跑走了 。\label{ex:1c}
\glt gou paozou le .
\glt `The dog(s) ran away.'

The word ``狗'' in~\ref{ex:1a} makes a general reference, translated as ``\emph{dogs}''. 
In the sentence~\ref{ex:1b}, the NP ``狗'' is indefinite, but whether it refers to a single dog or a set of dogs needs to be decided by wider contexts.
Likewise, the plurality status of the ``狗'' in the sentence~\ref{ex:1c} is hard to decide without further context, but it is certainly definite.

In this study, we build computational models to understand the following research question:
\begin{quote}
    \emph{To what extent plurality and definiteness of Chinese NPs are predictable from their contexts}?
\end{quote}
To this end, we formalised these two comprehension tasks as two classification tasks, i.e., classifying a Chinese NP as plural or singular and as definite or indefinite.
We first built a dataset, in which each NP is annotated with its plurality and definiteness, on the basis of a large-scale English-Chinese parallel corpus.
More specially, from the parallel corpus, we did word alignments and designed an algorithm to match NPs in the two languages based on the alignment results. We extracted NPs in Chinese and annotated the plurality and definiteness of each NP according to its matched English NP.
To guarantee the quality of the dataset, we conducted two human assessment studies which were then analysed and compared. 

We then performed a corpus analysis e.g. to investigate to what proportion plurality and definiteness are implicitly expressed. 
Subsequently, we tested mainstream classification techniques, from the classic machine learning based classifiers to the more recent pre-trained language model based classifiers, on the dataset to investigate the predictability of plurality and definiteness, and we analysed their behaviour.

\section{Dataset} \label{sec:data}

One of the major challenges of the present study is the construction of a large-scale dataset in which each NP is annotated with its plurality and definiteness. This is extraordinarily hard not only because building a large-scale human-annotated dataset is expensive, but also because many linguistic studies have demonstrated that deciding plurality and definiteness (especially definiteness) in Chinese NPs is a challenging task for even native speakers (e.g.,~\citet{robertson2000variability}).

Instead, inspired by~\citet{wang-etal-2016-novel}, in which they focused on pro-drop in machine translation systems, and the ``translation mining'' in corpus linguistics~\citep{bremmers2022translation}, since English speakers always convey plurality and definiteness explicitly, we can annotate a Chinese NP automatically if we have its English translation. Such information can be found in any English-Chinese parallel corpus.

More specifically, given a parallel corpus, we first did the word alignments and designed a simple but effective algorithm to extract and match NPs in both languages. Then, we annotated each Chinese NP based on its associated English NP. In what follows, we detail the automatic annotation process, introduce the resulting corpus and how we assess its quality.

\subsection{Dataset Construction}

Since we are investigating the pragmatics of Chinese NPs, the corpus needs to reflect the everyday use of language. In other words, the corpora that are constructed from news or novels are not appropriate. Therefore, we used the TV episode subtitle corpus, which was constructed and pre-processed by~\citet{wang2018translating}\footnote{The data came from two subtitle websites in China: \url{http://www.opensubtitles.org} and \url{http://weisheshou.com}.}. It contains 4.39 million English and Chinese sentence pairs in total. 

\paragraph{Word Alignment.} We used GIZA++~\citep{och-ney-2003-systematic} to generate alignment proposals. Note that the alignment proposal is sometimes different when aligning words in English to Chinese and when aligning words in Chinese to English. Therefore, at this step, we recorded the alignments of both ``directions'' for future use. 

\paragraph{NP Identification.} We used CoreNLP~\citep{manning-etal-2014-stanford} to parse each sentence in each language and extracted all NPs from the parse tree. We also recorded the Part-of-speech (POS) tag for each word in this step. 

\paragraph{NP Matching.} With the word alignments and identified NPs in hand, we design a simple but effective method in which there are two steps: (1) for each direction, an NP in the source language is paired with the NP in the target language that has the most aligned words with it; (2) a match is done if and only if two NPs are paired in both directions. 

\paragraph{Post-processing.} Since NPs are often in nested structures and not all NPs interested us, we filtered out some NPs: (1) we removed all NP conjunctions and only kept their constituents. For example, the NP ``\emph{Zhangsan and Lisi}'' contains two NPs. We remove it and keep only ``\emph{Zhangsan}'' and ``\emph{Lisi}''; (2) apart from NP conjunctions, for each NP, we dropped all its constituents. For example, if all of the ``\emph{Lisi's book}'', ``\emph{Lisi}'' and ``\emph{book}'' are matched in the previous step, we only keep ``\emph{Lisi's book}'' in our dataset. We also remove all NPs that are pronouns as they are not the focus of this study.  

\paragraph{Annotation.} For each Chinese NP, we annotated its matched English NP and used the resulting labels (i.e., plurality and definiteness) as its annotation. Concretely, we annotated an English NP as plural if: (1) it has a plural POS tag (i.e., \textsc{NNS} or \textsc{NNPS}); (2) it is a numeral phrase that specifies a quantity larger than one (i.e., ``\emph{two cups of coffee}''). Otherwise, it is a singular NP. For the definiteness of an NP, the annotation was done based on (1) its article; (2) whether it is a demonstrative phrase (i.e., whether it contains a demonstrative (decided based on its POS tag and its surface form), such as ``\emph{this}'' or ``\emph{that}''); and (3) whether it is a proper name (i.e., an NP is definite if it is a proper name).

\subsection{The Corpus}

\begin{table}
\small
\centering
\begin{tabular}{lcccc}
\toprule
  & \multicolumn{2}{c}{{\textsc{Plurality}}} & \multicolumn{2}{c}{{\textsc{Definiteness}}}\\
  \cmidrule(lr){2-3} \cmidrule(lr){4-5}
  & Singular & Plural & Definite & Indefinite \\ \midrule
  \textbf{train} & 79158 & 24528 & 48471 & 55215 \\
  \textbf{dev} & 7894 & 2474 & 4777 & 5591 \\
  \textbf{test} & 7925 & 2444 & 4844 & 5525 \\
 \bottomrule
\end{tabular}
\caption{The basic statistics of our dataset.}
\label{tab:statistics}
\end{table}

Due to the limitation of computing resources, we sampled and annotated 5\% of the data from~\citet{wang2018translating} for further computational modelling (described in the next section). Table~\ref{tab:statistics} charts the basic statistics of the resulting dataset. The dataset contains 124K annotated NPs. More than 3 quarters NPs are marked as singular. The definiteness labels are rather balanced. 58K samples are annotated as indefinite NPs while 66K samples are definite. We then divided the dataset into the training, development and test sets with the ratio 8:1:1.

\subsubsection{Is ``men'' a plural marker?} 

The inflectional morpheme ``们'' (men) was considered as a plural marker in Chinese. However, in the past several decades, theoretical linguists argued that it is indeed a collective marker \citep{iljic1994quantification, li1999plurality}, highlighting that a referent is a group of people and referring to the group as a whole (translated as ``\emph{group of}'' or ``\emph{set of}'') because it is incompatible with number phrase. Later on, \citet{huang2009syntax} further demonstrated that an NP with ``们'' \emph{must} be interpreted as definite. Therefore, it would be interesting to look into the labels of NPs with ``们'' in our dataset.

We extracted all NPs whose head noun has a ``们'' suffix\footnote{We remove NPs in which ``们'' does not function as suffixes, e.g., ``哥们'' (brother), and is part of pronouns, e.g., ``我们'' (we).} and did statistics on the label distribution. Regarding plurality, we found that although most extracted NPs were still marked as plural. There are still a remarkable amount of singular NPs (approximately, 9.12\%). This suggests that, in line with the linguistic theory, the suffix ``们'' is not a conclusive marker of plural. Regarding definiteness, inconsistent with what linguists suggested, most extracted NPs were marked as indefinite (approximately, 63.84\%). For example, the ``大人们'' (adults) in the example~\ref{ex:men} apparently does not have a definite reading. This embodies the conclusion that says ``们'' must be interpreted as definite is questioned.

\ex. 大人们 会 告诉 你 并 不是 这样 。\label{ex:men}
\glt darenmen hui gaosu ni bing bushi zheyang
\glt Adults will tell you this is not the case.

\subsubsection{How frequently do Chinese speakers express plurality or definiteness explicitly?}

For each NP in the dataset, we annotate whether it expresses plurality or definiteness explicitly based on the POS tags and the parsing tree of the sentence in which this NP is located. We marked an NP express plurality explicitly if it contains a numeral or a measure word. We marked an NP express definiteness explicitly if (1) it contains a proper name; (2) it includes a possessive; (3) there is a numeral or measure present, with a preceding demonstrative.

At length, we identified that merely 12.42\% utterances convey plurality explicitly and 15.86\% utterances contain explicit definiteness markers. This confirms that Chinese, as a ``cool'' language, its speakers indeed do not use explicit plurality and definiteness markers very often. 

\subsection{Quality Assessment}

Last but not least, it is essential to ensure that the corpus is suitable for 
use in computational modelling. We manually assess its quality from the aspects of plurality and definiteness annotation as well as NP identification. In what follows, we describe our assessment process.

\begin{table*}
\centering
\begin{tabular}{lcccccccc}
    \toprule
    & \multicolumn{4}{c}{{\textsc{Assessment 1}}} & \multicolumn{4}{c}{{\textsc{Assessment 2}}}\\
    \cmidrule(lr){2-5} \cmidrule(lr){6-9}
    & Acc$_{=2}$ & Acc$_{\geq 1}$ & IAA (\%) & IAA ($\kappa$) & Acc$_{=2}$ & Acc$_{\geq 1}$ & IAA (\%) & IAA ($\kappa$)\\ \midrule
    \textbf{NP Identification} & 79.50 & 96.25 & 0.8325 & - & - & - & - & - \\ 
    \textbf{Plurality} & 84.00 & 96.75 & 0.8725 & 0.6477 & 74.00 & 85.50 & 0.8850 & 0.6679 \\
    \textbf{Definiteness} & 81.00 & 97.25 & 0.8375 & 0.6731 & 53.00 & 77.50 & 0.7550  & 0.4755\\
    \bottomrule
\end{tabular}
\caption{Human Assessment Results, in which IAA (\%) is the percentage agreement and IAA ($\kappa$) is the Cohen's Kappa.}
\label{tab:human_assessment}
\end{table*}

\subsubsection{Assessment 1} 

We randomly sampled 400 samples for human assessment, the NP in each of which was highlighted. We hired four annotators and ensured that each sample was assessed by 2 annotators. All of them are native speakers of Chinese. Three of them are males and one of them is female. Two of them have backgrounds in engineering, one in statistics, and one in Language study.

Concretely, we asked annotators three questions (translated from Chinese): (1) Is the highlighted noun phrase correctly identified? (2) Is this a singular/plural (decided by the annotation in our corpus) phrase? and (3) Is this a definite/indefinite phrase? 

After the experiment, we computed the accuracy and inter-annotator agreements (IAA). We computed two types of accuracy based on the number of annotators in agreement with the annotation. Acc$_{=2}$ measures the proportion of accurate annotations agreed upon by both annotators, while Acc$_{\geq 1}$ measures those agreed upon by at least one annotator. For IAA, we computed both the percentage agreement and Cohen's Kappa~\citep{cohen1960coefficient}\footnote{We did not calculate Cohen's Kappa on raw human decisions because, for each question, the marginal probability of one answer is much greater than the other (i.e., there are much more ``yes'' than ``no''), which makes Cohen's Kappa inaccurate~\citep{brennan1981coefficient,maclure1987misinterpretation,donker1993interpretation}. 
Instead, we first translated their decisions in accordance with our labels. For example, if the label in our corpus is ``plural" and the annotator provided a positive response, we assumed that the annotator annotated this sample as ``plural".}.

Table~\ref{tab:human_assessment} charts the human assessment results. All three tasks received Acc$_{=2}$ around 80\% and Acc$_{\geq 1}$ higher than 96\%. One can ask why NP identification has received lower scores than the other two tasks. One major reason is that most identified incorrect NP identifications are about unsuccessfully including all modifiers (e.g., marking only ``\emph{the men}'' from the true NP ``\emph{the man who is old}'').

These results suggest that our corpus is of good quality, on the one hand. On the other hand, disagreements between two annotators exist in all three tasks. The percentage agreements of all three tasks are around 85\% and the Kappa values for the plurality and definiteness annotations are approximately 0.65, suggesting substantial agreements between annotators.
Nonetheless, we also noticed that the IAAs for the definiteness annotation are surprisingly high. This is counter-intuitive because, as aforementioned, many previous studies suggested that deciding the definiteness is hard for Chinese native speakers (e.g.,~\citet{robertson2000variability}). This may be attributed to the Framing Effects in human evaluation~\citep{schoch-etal-2020-problem}.  In particular, our use of yes or no questions might have influenced the evaluators' decisions, leading to a bias towards favouring a positive response. Additionally, such an influence may be magnified as disagreements exist by nature in our tasks. Therefore, we conducted assessment 2 as a complement.

\subsubsection{Assessment 2}

To minimise the bias introduced by the framing effects, in assessment 2, we gave each annotator samples from our dataset in which NPs were highlighted while labels were removed. We asked them to directly annotate the plurality and definiteness of each NP. This time, we sampled another 200 samples and, again, ensured that each sample is annotated by 2 annotators. The results are also reported in Table~\ref{tab:human_assessment}. 

Although Acc$_{=2}$ for plurality reduced from 84\% to 74\% while that for definiteness dramatically reduced from 81\% to 51\%, there are still approximately 80\% of our annotations agreed by at least one human annotator for both tasks. In this new assessment, the IAA for plurality stays high while the IAA for definiteness decreases. The kappa value for definiteness drops from 0.67 (assessment 1) to 0.48, indicating a moderate agreement. These results cohere with what linguists suggested. 

\subsubsection{Summary}

We found that, for all three tasks, disagreements (between two annotators and between the annotators' annotation and our annotation) exist and differ with respect to how the questions are framed. Despite the disagreements, the assessment results indicate that our corpus is of acceptable quality. In both assessments, at worst, approximately 80\% of our annotations can be agreed upon by at least one human annotator.

\subsection{Limitations} \label{sec:limitation}

Regarding our annotation and assessment processes, our corpus exhibits the following limitations: \textbf{First}, as shown in the assessment experiments, disagreements exist in the human annotation. This is true for many pragmatic tasks~\citep{poesio-etal-2019-crowdsourced}. However, our automatic annotation strategy cannot take such agreements into consideration. \textbf{Second}, Chinese does not distinguish countable and uncountable nouns. By looking into the human annotation results from assessment 2, we found that since Chinese is a classifier language (i.e., numerals obligatorily appear with classifiers when they modify nouns) many NPs with uncountable nouns were considered plural NPs. Because we annotate NPs in Chinese using the information from their English translations, we failed to annotate these uncountable nouns correctly. 
\textbf{Thrid}, both our automatic annotation and human assessments are precision-oriented. For example, we dropped the Chinese NP that did not match with any English NPs and, during the assessments, we only used NPs that had been matched. This makes our corpus overlook some Chinese NPs and our assessments ignore recall. \textbf{Last}, in the assessments, we did not evaluate how the decisions of annotators would be influenced by providing them with additional contexts for each sample. This limitation was recognised because, as mentioned in Section~\ref{sec:intro}, the meaning of a Chinese NP relies more on its context compared to its English counterpart.

\section{Models} \label{sec:model}

In this section, we introduce models we built for predicting plurality and definiteness. We tried a large variety of models: from classic machine learning (ML) based models to the most recent pre-trained language model (PLM) based models. 

\subsection{ML-based Models}

We tried a number of classic ML-based classifiers on our plurality and definition prediction tasks. To this end, we first used `*' to mark the target NP in each sample. For example, ``我 的 母亲'' (my mom) is the target NP in the following sentence. 
\ex. 我 爱 * 我 的 母亲 * 。
\glt wo ai * wo de muqin * .
\glt I love * my mom * .

We used N-gram ($N=1,2,3,4$) as features for classification\footnote{The performance of our classifiers can be further boosted using advanced features, e.g., POS tags or syntactic structures. Since, in this study, we were investigating the predictability of plurality and definiteness of NPs from their contexts, we used only raw features from the contexts.}. As for the algorithm, we tried Random Forest (RF), Logistic Regression (LR) and Support Vector Machine (SVM).

\subsection{PLM-based Models}

\begin{figure}
    \centering
    \includegraphics[scale=0.45]{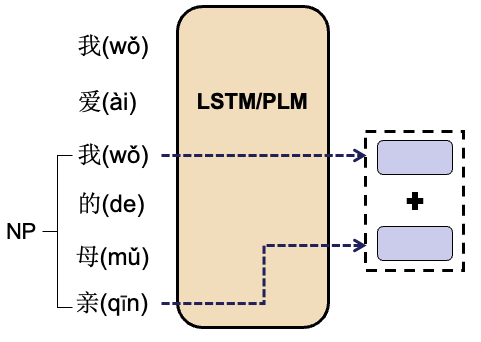}
    \caption{Illustration of the PLM-based Models.}
    \label{fig:model}
\end{figure}

\begin{table*}[t!]
\label{plurality model performance}
\centering
\small
\begin{tabular}{lcccccccccccc}
\toprule
& \multicolumn{6}{c}{Plurality} & \multicolumn{6}{c}{Definiteness}\\ 
\cmidrule(lr){2-7} \cmidrule(lr){8-13}
  & \multicolumn{3}{c}{{\textsc{Macro avg}}} & \multicolumn{3}{c}{{\textsc{Weighted avg}}} & \multicolumn{3}{c}{{\textsc{Macro avg}}} & \multicolumn{3}{c}{{\textsc{Weighted avg}}} \\
  \cmidrule(lr){2-4} \cmidrule(lr){5-7} \cmidrule(lr){8-10} \cmidrule(lr){11-13}
  & P & R & F & P & R & F & P & R & F & P & R & F \\
  \midrule
  RF & 81.08 & 58.19 & 58.53 & 80.26 & 79.69 & 74.19 & 68.63 & 67.24 & 67.10 & 68.51 & 68.09 & 67.47 \\
  LR & 76.08 & 67.39 & 69.79 & 80.11 & 81.58 & 79.77 & 71.73 & 71.53 & 71.58 & 71.78 & 71.82 & 71.75 \\
  SVM & 75.56 & 67.37 & 69.69 & 79.88 & 81.40 & 79.65 & 71.34 & 71.04 & 71.10 & 71.37 & 71.40 & 71.29 \\\midrule
  \bilstm & 79.31 & 70.94 & 73.59 & 82.49 & 83.50 & 82.14 & 76.78 & 76.88 & 76.80 & 76.95 & 76.84 & 76.87 \\
  \bert & 80.88 & \underline{77.96} & \underline{79.24} & \underline{85.23} & 85.73 & 85.37 & 81.60 & 81.66 & 81.63 & 81.71 & 81.69 & 81.69 \\
  \bertwwm & 80.94 & \textbf{78.34} & \textbf{79.50} & \textbf{85.38} & \underline{85.83} & \textbf{85.52} & \underline{81.95} & \underline{81.82} & \underline{81.87} & \underline{81.98} & \underline{81.98} & \underline{81.97} \\
  \mbert & 80.07 & 76.96 & 78.30 & 84.58 & 85.15 & 84.74 & 80.70 & 80.41 & 80.50 & 80.68 & 80.66 & 80.62 \\
  \roberta & \underline{81.21} & 77.53 & 79.09 & 85.22 & 85.79 & 85.35 & \textbf{82.27} & \textbf{82.10} & \textbf{82.16} & \textbf{82.28} & \textbf{82.28} & \textbf{82.26} \\
  \roberta & \textbf{81.72} & 77.37 & 79.17 & \textbf{85.38} & \textbf{85.98} & \underline{85.46} & 81.80 & 81.58 & 81.66 & 81.79 & 81.79 & 81.76 \\
  (large) & & & & & & & & & & & & \\
\bottomrule
\end{tabular} 
\caption{The performance of our models for plurality and definiteness predictions depicted in Section~\ref{sec:model}. ``P'', ``R'' and ``F'' stand for precision, recall and F-score respectively. The best results are \textbf{boldfaced}, whereas the second best are \underline{underlined}. The PLMs that do not mark `(large)' use their base version. For many Chinese PLMs, only the base models are publicly available.}
\label{tab:results}
\end{table*}

Recently, the developments in NLP to a large extent attributed to the introduction of PLMs. This contribution stems from two perspectives: utilising the knowledge acquired through large-scale pre-training and leveraging a broader context. Recall that we are investigating whether the plurality and definiteness of an NP can be predicted from its context. Therefore, it is plausible to assume that such predictions also benefit from using (contextual) PLMs. 

To this end, we fine-tune PLMs on our dataset. As depicted in Figure~\ref{fig:model}, we fed the raw text into a PLM, and, for each NP, we extracted the representations of its first token and its last token. The prediction was made by a dense output layer based on the summation of these two representations. 

In this study, we tried the following PLMs: (1) Chinese \bert and \roberta~\citep{devlin-etal-2019-bert, liu2019roberta}. (2) \bertwwm: vanilla Chinese BERT was pre-trained as a fully character-based model, but \citet{cui2021pre} proved that the performance can be boosted if Whole Word Masking (WWM; rather than character level masking) is done during pre-training. (3) \mbert: since in addition to Chinese, there are multiple other ``cool'' languages (e.g., Japanese, Korean and Arabic), we, therefore, wanted to validate whether the predictions can benefit from multilingual pre-training or not. (4) \bilstm: In addition to PLMs, we also tested bi-directional LSTM~\citep{schuster1997bidirectional} initialised by the Glove embeddings~\citep{pennington-etal-2014-glove}. The architecture is the same as how we used PLMs.
\section{Experiments} \label{sec:experiment}

In this section, we introduce the evaluation protocol and report the performance of the models.

\subsection{Evaluation Protocol}

We tuned the hyper-parameters of each of our models on the development set and chose the setting with the best macro F1 score. We report the macro/weighted averaged precision, recall, and F1 on the test set.

\subsection{Experimental Results}

Table~\ref{tab:results} depicts the results of the plurality and definiteness classifications. The results suggest that all models can learn useful information for both plurality and definiteness predictions. 
Similar to human beings, models also face more challenges when making predictions about definiteness compared to plurality, as evidenced by the lower weighted scores in definiteness predictions compared to plurality predictions.

For model performance, as expected, PLM-based models outperformed their ML-based counterparts. Among ML-based models, we found that LR is very effective, achieving weighted-averaged F-scores of 79.77 for plurality predictions and 71.75 for definiteness predictions. \bilstm with Glove embeddings defeated all ML-based models but lost to Models with \bert. This embodies that context plays an important role in the prediction of plurality and definiteness, which is consistent with the definition of ``cool'' (see Section~\ref{sec:intro}).

Among \bert-based models, we had the following observations: (1) \bertwwm performed remarkably well. It generally performed the best for plurality prediction and was the second-best model for definiteness prediction. This demonstrated that, on pragmatics tasks (e.g., our tasks), \bert does benefit from whole word mask pre-training probably because the intended meaning of a word (noun in our situation) is mainly inferred from its context rather than its inner structure. (2) \bert did not benefit from multilingual pre-training as \mbert received 84.74 weighted F-score on plurality predictions and 80.62 on definiteness predictions though \mbert was pre-trained on typical ``cool'' languages, including Arabic, Japanese and Korean. This is probably attributed to the fact that speakers of these ``cool'' languages use contexts differently and, therefore, multi-lingual pre-training may not yield substantial benefits to downstream tasks that rely on context.
This makes supervision signals become needed. In the future, it would be valuable to build an NP corpus in multiple ``cool'' languages and see whether the predictions can benefit or not. (3) Interestingly, on our tasks, the amount of parameters is not the more the better. \roberta-large performed worse than the vanilla \bert on plurality predictions and worse than \roberta-base on definiteness predictions. Further probing experiments are needed to explain what happens. 
\section{Analysis}

In what follows, we analyse the model behaviour concerning three questions.

\begin{table*}[t]
\centering
\small
\begin{tabular}{lcccccccccccc}
\toprule
& \multicolumn{6}{c}{4-way} & \multicolumn{6}{c}{2-way (merged)}\\ 
\cmidrule(lr){2-7} \cmidrule(lr){8-13}
  & \multicolumn{3}{c}{{\textsc{Macro avg}}} & \multicolumn{3}{c}{{\textsc{Weighted avg}}} & \multicolumn{3}{c}{{\textsc{Macro avg}}} & \multicolumn{3}{c}{{\textsc{Weighted avg}}} \\
  \cmidrule(lr){2-4} \cmidrule(lr){5-7} \cmidrule(lr){8-10} \cmidrule(lr){11-13}
  & P & R & F & P & R & F & P & R & F & P & R & F \\ \midrule
  \bert & 67.37 & 64.26 & 65.53 & 70.72 & 71.20 & 70.79 & 65.62 & 63.35 & 64.34 & 69.49 & 69.91 & 69.61 \\
  \bertwwm & 67.94 & \underline{65.74} & 66.72 & 71.54 & 71.86 & 71.62 & 66.51 & \textbf{64.23} & \textbf{65.24} & \underline{70.03} & \underline{70.40} & \underline{70.14}\\
  \mbert & 67.73 & 64.58 & 65.69 & 71.12 & 71.46 & 71.01 & 64.19 & 61.51 & 62.62 & 68.11 & 68.59 & 68.21 \\
  \roberta & \underline{68.25} & \textbf{66.42} & \textbf{67.24} & \underline{72.03} & \underline{72.36} & \underline{72.14} & \underline{67.08} & \underline{63.89} & \underline{65.23} & \textbf{70.29} & \textbf{70.74} & \textbf{70.36} \\
  \roberta & \textbf{68.73} & 65.51 & \underline{66.87} & \textbf{72.09} & \textbf{72.55} & \textbf{72.18} & \textbf{67.11} & 63.36	& 64.90 & 69.90 & 70.35 & 69.92 \\
  (large) & & & & & & & & & & & & \\
\hline
\end{tabular}
\caption{The results of 4-way prediction and the merged results of 2 binary predictions.}
\label{tab:4way}
\end{table*}

\subsection{What is the impact of Context Size?}

\begin{figure}
    \centering
    \includegraphics[scale=0.55]{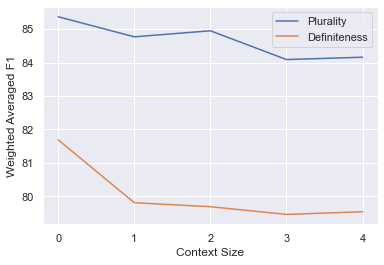}
    \caption{Weighted F1 concerning different context sizes. The size is measured by the number of sentences around the target sentence.}
    \label{fig:context}
\end{figure}

According to what \citet{huang1984distribution} hypothesised, the interpretation of the plurality and definiteness of an NP relies on its context and such context is not necessarily only the current sentence but also the whole discourse. For example, without more context, it is hard to decide the plurality of the NP in example~\ref{ex:1c}. However, in the current experimental setting, we only fed the models with only one sentence, namely the target sentence. 

Therefore, it is plausible to expect that if we increase the size of contexts, the predictions become more accurate. To validate this idea, we increased the size and assessed \bert with the inputs with different amounts of contexts. Figure~\ref{fig:context} prints the evaluation results of the two tasks, in which 1 means both the previous sentence and the preceding sentence are seen as the context and be fed to the models together with the target sentence.

Nevertheless, different from the expectation, the performance of both tasks decreases with the increase of the context size. The decrease in performance is more pronounced in definiteness prediction compared to plurality prediction. A possible explanation is that although wider contexts add useful information to the prediction, it also adds confusion as our focus is only a small part (i.e., the NP) of the target sentence. This makes it hard for the model to extract useful information from the representation of a wide context, and add it to the representation of the target NP (which is often a few words; recall that we only used the representation of the target NP for prediction), and make predictions. It is worth noting that similar phenomena are observed in other pragmatics tasks~\citep{joshi-etal-2019-bert,baruah-etal-2020-context,same-etal-2022-non,chen2023neural}.

\subsection{Do the plurality and definiteness predictions help each other?}

\begin{figure}
    \centering
    \includegraphics[scale=0.55]{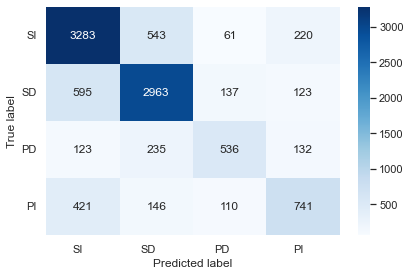}
    \caption{The confusion matrix for 4-way prediction of \roberta-large, in which S, P, I and D mean ``singular'', ``plural'', ``indefinite'' and ``definite'', respectively.}
    \label{fig:confision}
\end{figure}

\begin{figure*}
    \centering
    \includegraphics[scale=0.65]{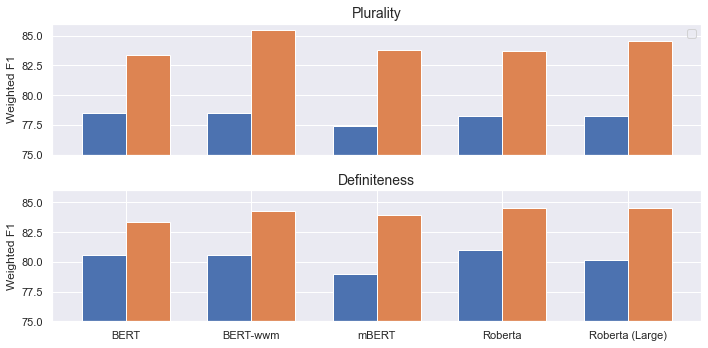}
    \caption{Macro F-scores of \bert-based models on implicit and explicit expressions of plurality and definiteness. The blue bars indicate the performance of models on implicit expressions while the orange bars indicate that on explicit expressions.}
    \label{fig:implicit}
\end{figure*}

Since both plurality and definiteness are information carried by NPs. One could expect that the information that is needed for predicting the plurality of an NP might help determine the definiteness of the same NP and vice versa. In other words, we might benefit from predicting plurality and definiteness simultaneously. Rather than employing multi-task learning, we opted to fine-tune the models for 4-way predictions. Specifically, given an NP, the models classify it into one of four categories: indefinite singular, indefinite plural, definite singular, or definite plural. To fairly compare the model performance for 4-way prediction and 2 separate binary predictions, we merged the predictions obtained in Section~\ref{sec:experiment} and re-computed each score. 

Table~\ref{tab:4way} reports the performance of each model on 4-way and merged binary predictions. The results suggest that models can significantly benefit from predicting plurality and definiteness simultaneously compared to predicting them separately. For example, in joint prediction, \roberta achieved a weighted average F1 score of 72.14. However, when doing binary predictions, the merged weighted F1 score dropped to 70.36.

Focusing on the 4-way prediction results, we found that akin to the binary predictions, \roberta had the best performance. It achieved a weighted F1 score of 72.14 and a micro F1 score of 67.24. It was followed by \roberta-large, who had an on-par weighted F1 and lower micro F1. \bertwwm performed slightly worse than them, but still remarkably well. Figure~\ref{fig:confision} is the confusion matrix of Roberta-large for joint prediction, which further ascertains the theory that deciding definiteness is hard in Chinese as although the labels of the plurality are way more imbalanced than that of the definiteness (see Table~\ref{tab:statistics}) the model is still much easier to confuse between ``definite'' and ``indefinite'' than between ``singular'' and ``plural''.

\subsection{How does the explicitness impact the model's behaviours?}

In the corpus analysis, we identified that NPs in 12.42\% and 15.86\% samples from our dataset explicitly express plurality and definiteness respectively. Since these explicit expressions provide clear markers, we expected that the predictions of both tasks on explicit expressions are easier than on implicit expressions. Thus, models would receive higher scores on the portion of explicit expressions. To examine this, we assessed \bert-based models on implicit and explicit expressions respectively and report the results in Figure~\ref{fig:implicit}\footnote{To highlight the differences, we report macro-F this time.}. As expected, for both tasks, all models performed better on explicit expressions and implicit expressions. 

Besides, we also have some interesting observations: (1) the difference between the performance on explicit expressions and on implicit expressions is larger on plurality prediction than definiteness prediction. (2) For plurality prediction, except \mbert, all other models have similar performance on implicit expressions. \bertwwm performed significantly better on explicit expressions than other models. (3) For definiteness prediction, \roberta performed the best on both implicit and explicit expressions.

\section{Conclusion}

We investigated one pragmatic aspect of the ``coolness'' hypothesis by~\citet{huang1984distribution}: in a ``cool'' language, whether the meaning of an omittable component is predictable or not.
To this end, we studied the predictability of plurality and definiteness in Chinese NPs, which, syntactically, are omittable.
We first constructed a Chinese corpus where each NP is marked with its plurality and definiteness. 
Two assessment studies showed that our corpus is of good quality. 
A corpus analysis suggests that Chinese speakers frequently drop plural and definiteness markers. 

Based on the corpus, we built computational models using both classic ML-based models and the most recent PLM-based models. The experimental results showed that both ML-based models and PLM-based models can learn information for predicting the meaning of plurality and definiteness of NPs from their contexts and that \bertwwm generally performed the best due to its good ability to extract information from contexts in Chinese. Further analyses of the models suggested that the information for predicting plurality and definiteness benefits from each other.

Regarding ``coolness'', through computational modelling, we confirmed that the plurality and definiteness of Chinese NPs are predictable from their contexts.
Furthermore, these predictions can be improved if the model's ability to capture contexts is enhanced.
Nonetheless, in addition to the research question presented in the current study (see Section~\ref{sec:intro}), another crucial question remains unanswered: to what extent do these computational models mimic listeners' way of comprehending plurality and definiteness? To address this question in the future, we intend to create a corpus in which disagreements among listeners are annotated, which is then used for assessing computational models.

\section{Ethical Considerations and Limitations}

In this work, potential biases may have two sources. One is the dataset. Our dataset was built from TV episodes, which have never been filtered with respect to toxic content. The other is the pre-trained language models we used, which have been widely discussed in, e.g.,~\citet{bender2021dangers}.

In addition to the limitations we discussed in Section~\ref{sec:limitation}, another key limitation is that our corpus analyses and computational modelling were done on the data from a single source, namely, conversations in TV episodes. It is not fully clear whether our findings can be generalised to data in other genres. Moreover, the data we used is a parallel corpus, where the Chinese texts were translated from English. While these texts maintain a natural tone, there's a risk that translations may diverge from everyday language use.

\nocite{*}
\section{Bibliographical References}\label{sec:reference}

\bibliographystyle{lrec-coling2024-natbib}
\bibliography{custom}

\end{CJK}
\end{document}